\documentclass[letterpaper, 10 pt, conference]{ieeeconf}  

\usepackage[pdftex]{graphicx}
\usepackage{amsmath}
\usepackage{subcaption}
\usepackage{tikz}
\usetikzlibrary{decorations.markings, calligraphy}
\usepackage{pgfplots}
\usepackage{url}

\IEEEoverridecommandlockouts                              

\title{\LARGE \bf
Removing Adverse Volumetric Effects From Trained Neural Radiance Fields
}

\author{Andreas L. Teigen$^{1*}$ \hspace*{1cm} Mauhing Yip$^{2*}$ \hspace*{1cm} Victor P. Hamran$^{1}$  \hspace*{1cm} Vegard Skui$^{1}$ \\ Annette Stahl$^{2}$ \hspace*{1cm} Rudolf Mester$^{1}$
\thanks{$^{1}$Authors are with the Department of Computer Science, Norwegian University of Science and Technology}%
\thanks{$^{2}$Authors are with the Department of Engineering Cybernetics, Norwegian University of Science and Technology}%
\thanks{$^{*}$Equal contribution.}
\thanks{For questions, send email to: {\tt\small andreas.l.teigen@ntnu.no}}%
\thanks{This paper is financially supported by the Norwegian Research Council in the project Autonomous Robots for Ocean Sustainability (AROS),
project number 304667.}
}

\begin{document}

\maketitle
\thispagestyle{empty}
\pagestyle{empty}

\begin{abstract}

While the use of neural radiance fields (NeRFs) in different challenging settings has been explored, only very recently have there been any contributions that focus on the use of NeRF in foggy environments. We argue that the traditional NeRF models are able to replicate scenes filled with fog and propose a method to remove the fog when synthesizing novel views. By calculating the global contrast of a scene, we can estimate a density threshold that, when applied, removes all visible fog. This makes it possible to use NeRF as a way of rendering clear views of objects of interest located in fog-filled environments. Additionally, to benchmark performance on such scenes, we introduce a new dataset that expands some of the original synthetic NeRF scenes through the addition of fog and natural environments. The code, dataset, and video results can be found on our project page: https://vegardskui.com/fognerf/

\end{abstract}

\section{INTRODUCTION}

\label{sec:intro}
Advances in NeRFs have allowed for the creation of photo-realistic reconstructions and novel view synthesis of real-world scenes given a set of camera images with known poses. It has quickly been adopted by the robotics community to aid in tasks like localization \cite{maggio2023loc}, mapping \cite{sucar2021imap}, and object manipulation \cite{dai2023graspnerf}. Although NeRF achieves state-of-the-art results at tasks in clear environments, little attention has been given to addressing other real-world adverse conditions like fog and haze. We address the task of extracting clear-view images of the solid objects of interest after training NeRF models in such adverse conditions.

\begin{figure}
    \centering
    \includegraphics[width=0.49\linewidth]{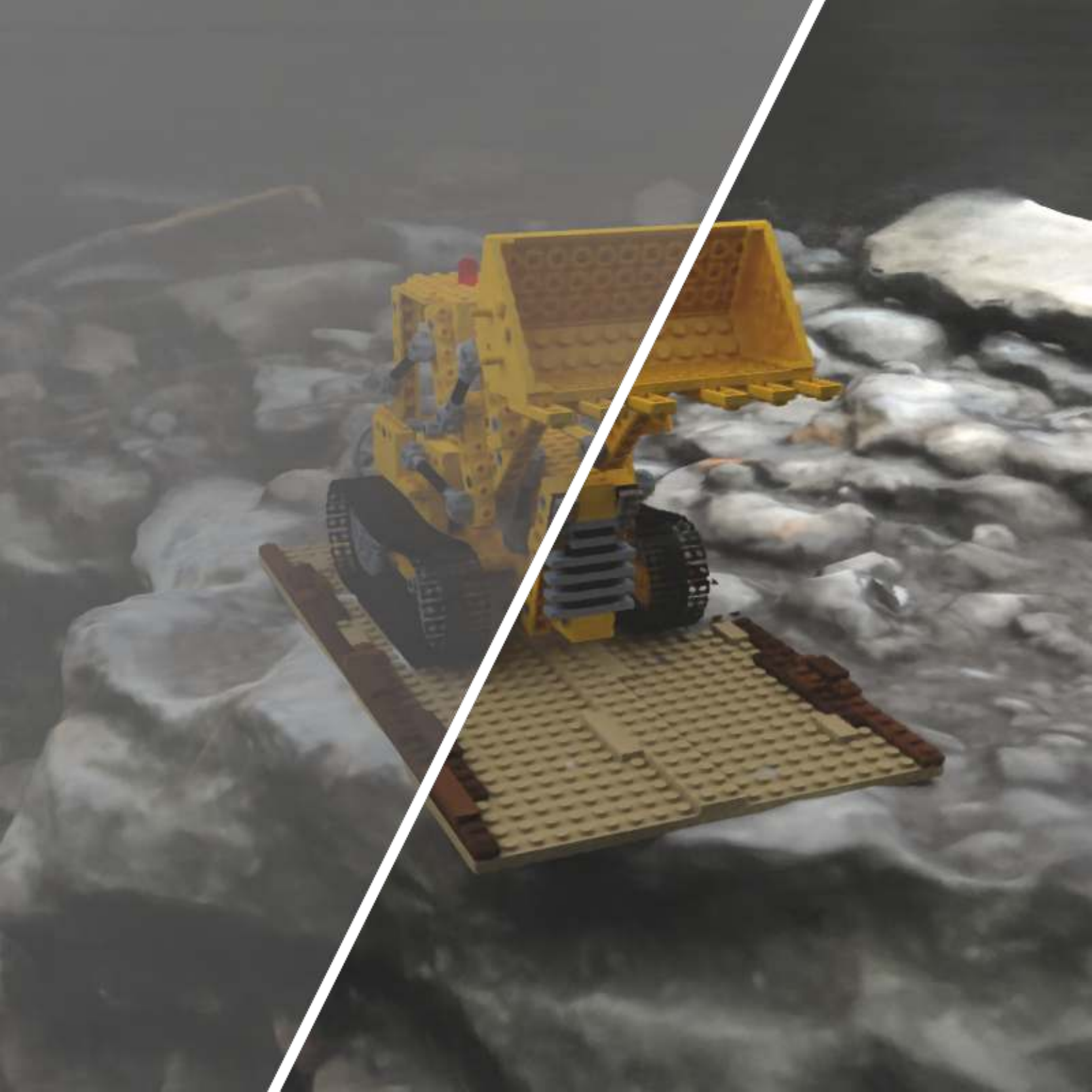}\hspace{1mm}%
    \includegraphics[width=0.49\linewidth]{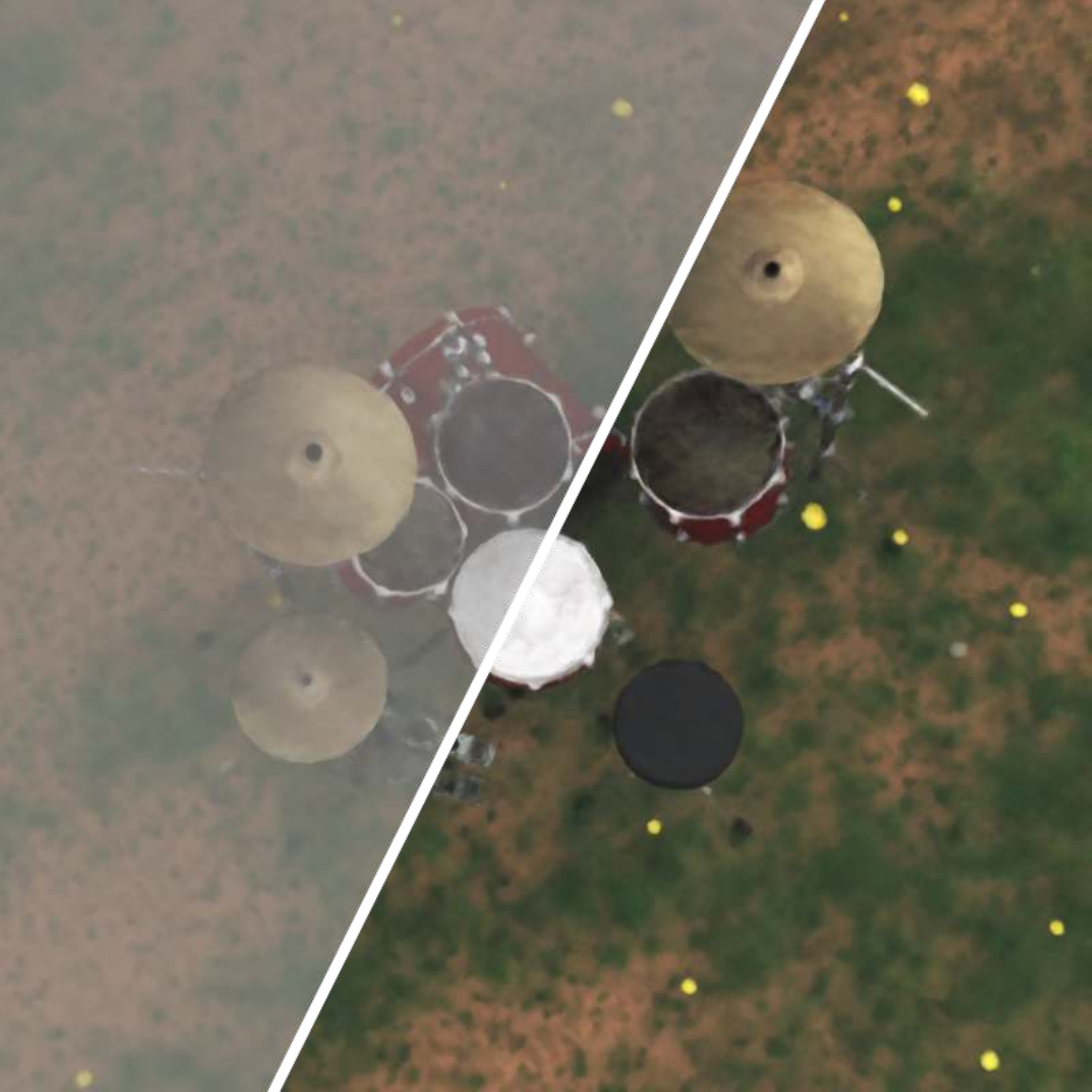}\\
    \vspace{1mm}%
    \includegraphics[width=0.49\linewidth]{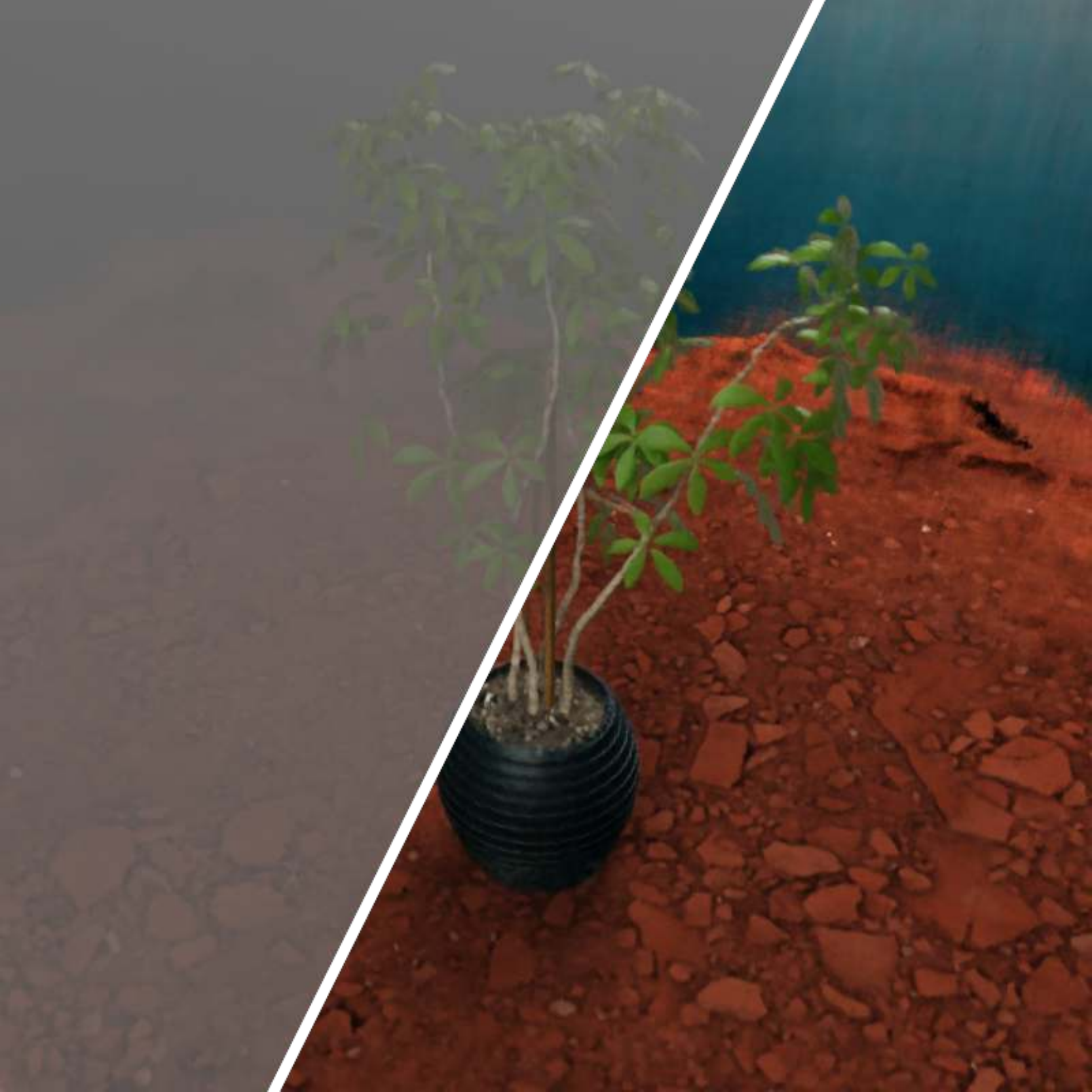}\hspace{1mm}%
    \includegraphics[width=0.49\linewidth]{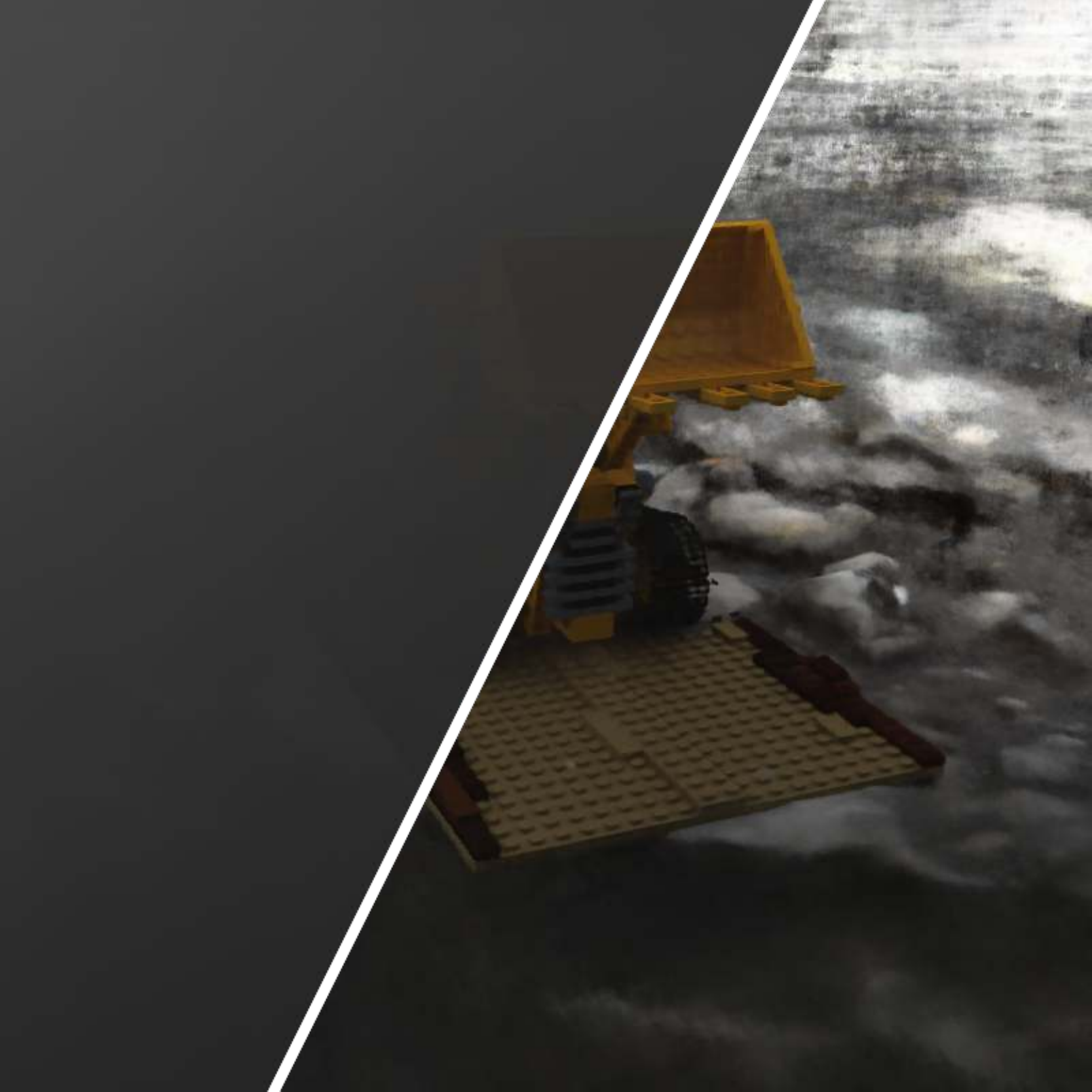}
    \caption{Generic NeRF model trained on hazy data, before (top left) and after (bottom right) applying our algorithm.}
    \label{fig:ad}
\end{figure}


In this paper, we propose a general method of rendering fog-free images from radiance fields trained on foggy data. Our method extends the post-training rendering pipeline and can be used with any generic, pre-trained NeRF model. Exploiting the ability of NeRF to learn densities as an implicit function across the whole volume of a scene, we show that simple density thresholding is enough to extract fog-free images from the radiance field while preserving the finer details of the model. Assisting this process, we propose a method to automatically estimate such a threshold after a model has been trained to convergence. Using our approach, this estimation only needs to be completed once for any pre-trained model. By requiring approximately the same computation as rendering a single frame, the process results in a negligible time loss when rendering multiple images. Some results are shown in fig. \ref{fig:ad}.

To assess the performance of NeRF and other novel view synthesis/reconstruction methods in foggy environments, we introduce a dataset with four new synthetic scenes. We invite other researchers to use our dataset, which can be downloaded from our project page.

In summary, our contributions are as follows:
\begin{enumerate}
    \item Demonstrate that volumetric fog is well captured by a standard NeRF model.
    \item Show that this fog can be removed during view synthesis with a simple threshold on density.
    \item Create an algorithm capable of automatically determining the value for which to threshold the density to remove this fog.
    \item Produce and publish a dataset for novel view synthesis augmented with fog effects.
\end{enumerate}

\section{Related Work}
\label{sec:relatedwork}

\paragraph{Neural Radiance Fields for View Synthesis.} 
The field of 3D computer vision has been greatly affected by the popularization of deep neural networks as universal function approximations \cite{hornik1989multilayer}. The task of synthesizing new views of a scene based on a set of posed images is one of the branches in computer vision that has noticed this effect through the introduction of Neural Radiance Fields \cite{mildenhall2020nerf}. NeRFs use a multilayer perceptron (MLP) as the sole representation of a scene by approximating a radiance field and synthesizing new, highly detailed novel views through differentiable volume rendering techniques. The radiance field, i.e., the MLP, is optimized to generate a volumetric model that can render replicas of posed images through stochastic gradient descent based on photometric reconstruction loss for individual pixels. There have been numerous extensions to NeRFs with regard to; scalability by unbounding or combining the radiance fields \cite{Barron_2022_CVPR, zhang2020nerf++, tancik2022block}, anti-aliasing through conical frustums instead of rays~\cite{Barron_2021_ICCV}, better specular reflections and relighting~\cite{verbin2022ref, bi2020neural, Boss_2021_ICCV, Srinivasan_2021_CVPR}, very sparse sets of images~\cite{niemeyer2022regnerf}, and representing dynamic scenes~\cite{Pumarola_2021_CVPR, Gao_2021_ICCV, attal2023hyperreel}. Furthermore, the use of depth in RGB-D images to further improve the accuracy and convergence time has been explored~\cite{Sucar_2021_ICCV, zhu2022nice, Johari_2022_CVPR, Azinovic_2022_CVPR, dey2022mip, stelzner2021decomposing}. We believe our method can extend all these methods due to the simplicity of our proposal. That said, we follow the original implementation of NeRF~\cite{mildenhall2020nerf} to keep simplicity low.

\paragraph{NeRFs in Challenging Settings.}
There has been significant research done to make NeRF applicable to different settings. First of all, NeRF-W~\cite{Martin-Brualla_2021_CVPR} extends NeRF to work on unconstrained photo collections, e.g. variable illumination and transient occluders, allowing for accurate reconstructions from unstructured image collections taken from the internet. Urban radiance fields~(URF)~\cite{Rematas_2022_CVPR} builds upon NeRF-W through the addition of LIDAR sweeps to reconstruct urban environments such as data from Google Street View~\cite{streetview}. URF uses a semantic segmenter to remove people or varying sky from images, this requires an oracle that detects outliers for arbitrary distractors. In response, RobustNeRF~\cite{sabour2023robustnerf} avoids this through the use of \textit{robust loss}, i.e., loss based on photometrically-inconsistent observations, without a priori knowledge of the types of distractors. Furthermore, NeRF in the Dark~\cite{Mildenhall_2022_CVPR} aims to train on linear raw images from scenes in dark environments, preserving the scene's full dynamic range, thus allowing for novel high dynamic range~(HDR) view synthesis. 

In parallel to our work, there have recently been published multiple contributions that aim to represent hazy scenes and render novel views using NeRFs. It was shown in \cite{li2022climatenerf} that fog can be added to a trained NeRF model by simply adding a non-negative constant to the density in the radiance field. The robustness of NeRF to adverse effects, including fog, was explored in \cite{wang2023benchmarking}, where they model 3D aware fog based on \cite{Kar_2022_CVPR} for synthetic data but have not made this dataset publicly available. They also add fog to real data, but the method is not depth-aware, nor is it consistent across images, making it an unrealistic scenario. 

Several works like \cite{sethuraman2022waternerf, zhang2023beyond, levy2023seathru, chen2023dehazenerf, li2023dehazing} tries to remove volumetric effects like fog \cite{chen2023dehazenerf, li2023dehazing} or an underwater medium \cite{sethuraman2022waternerf, zhang2023beyond, levy2023seathru} using a NeRF model. Furthermore, \cite{sethuraman2022waternerf, zhang2023beyond} focuses on color restoration of the scene as if there were no volumetric medium, while \cite{chen2023dehazenerf, li2023dehazing} prioritizes the removal of the medium. The latter will result in a somewhat darker image due to the underlying model being partially occluded due to the haze. Decidedly different from our method is how these methods require explicit modeling of volumetric effects using custom architectural variations on the NeRF model, while our method works for any pre-trained model.

Concurrently and independently of our work, \cite{jin2023reliable} presents an algorithm for fog removal on pre-trained NeRF models. The method differs from ours by determining the threshold using changes between images at certain intervals instead of the contrast metric we use. Furthermore, neither the code nor the dataset has yet been published for comparison.

\begin{figure*}
    \vspace{2mm}
    \centering
    
    \rotatebox[origin=c]{90}{Blender}\hspace{2mm}%
    \begin{subfigure}[]{0.235\textwidth}
        \centering
        \includegraphics[width=0.48\textwidth]{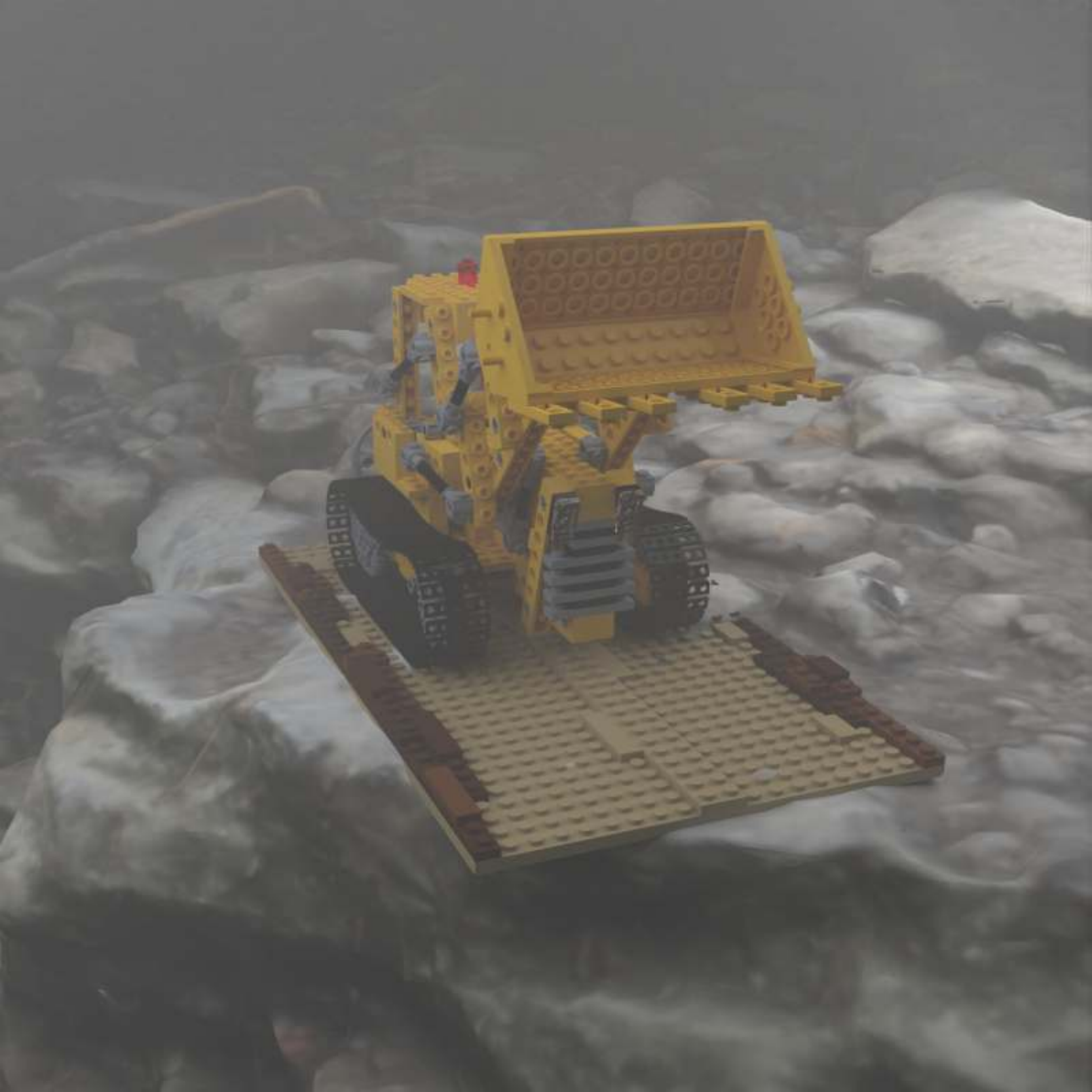}\hspace{1mm}%
        \includegraphics[width=0.48\textwidth]{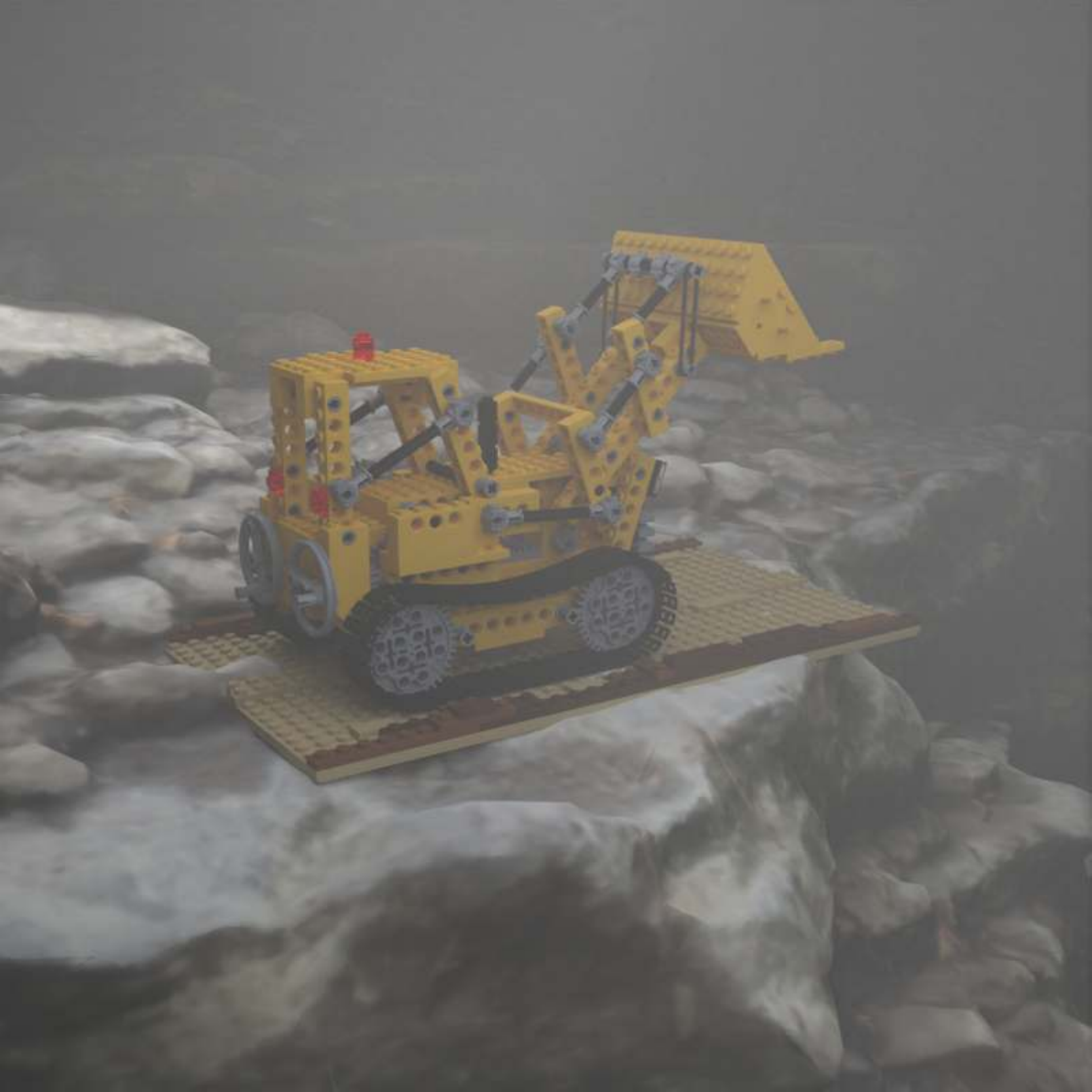}
    \end{subfigure}\hspace{1mm}%
    \begin{subfigure}[]{0.235\textwidth}
        \centering
        \includegraphics[width=0.48\textwidth]{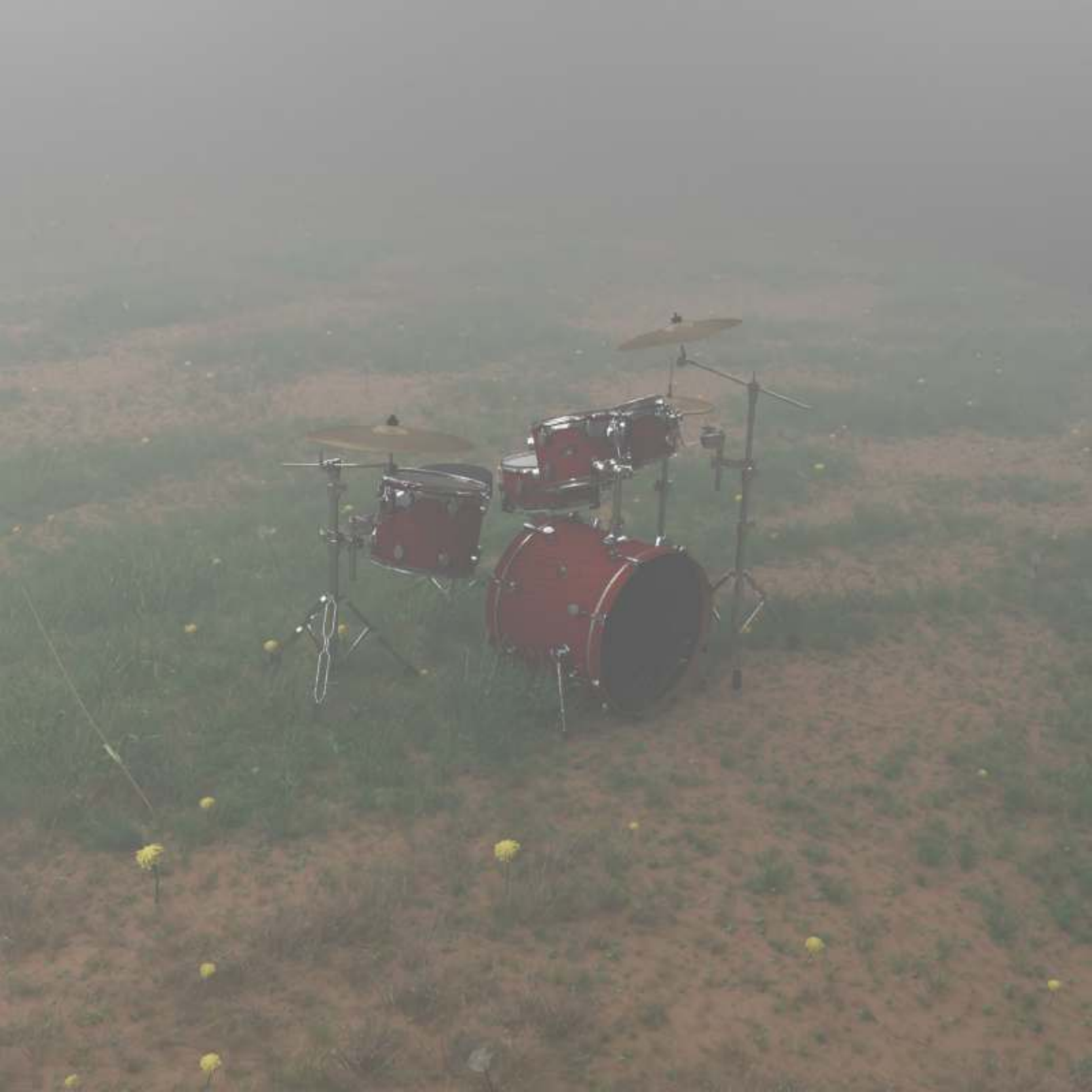}\hspace{1mm}%
        \includegraphics[width=0.48\textwidth]{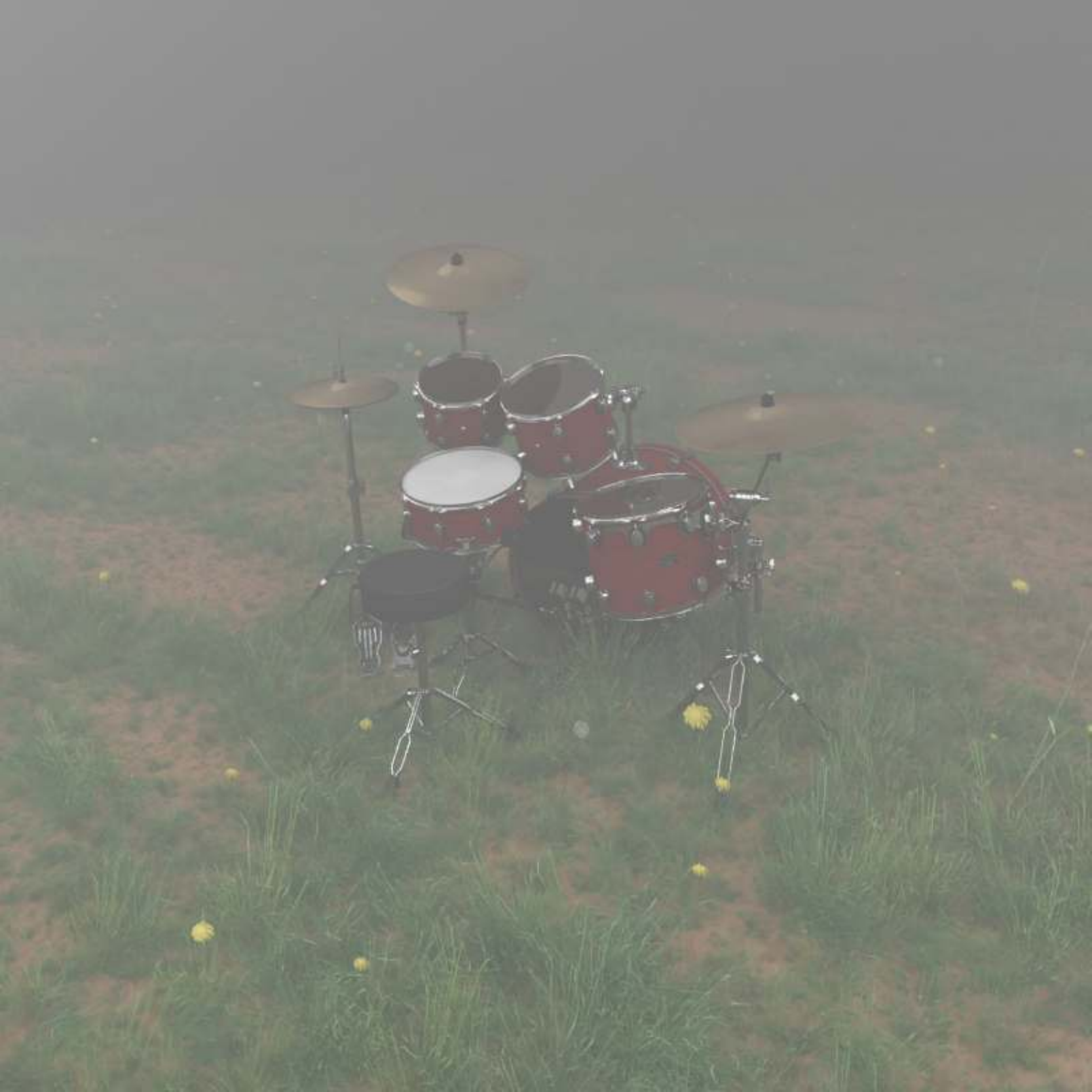}
    \end{subfigure}\hspace{1mm}%
    \begin{subfigure}[]{0.235\textwidth}
        \centering
        \includegraphics[width=0.48\textwidth]{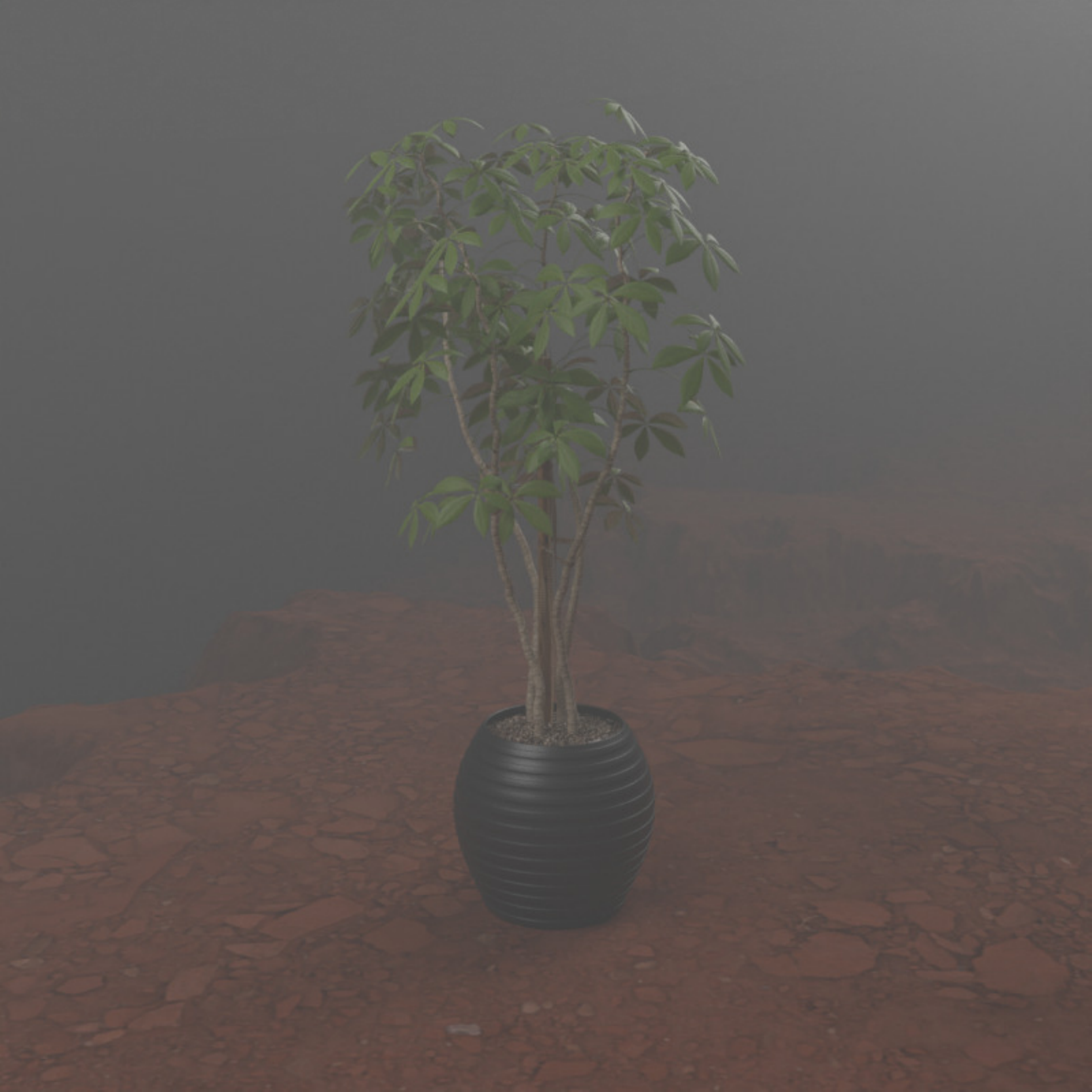}\hspace{1mm}%
        \includegraphics[width=0.48\textwidth]{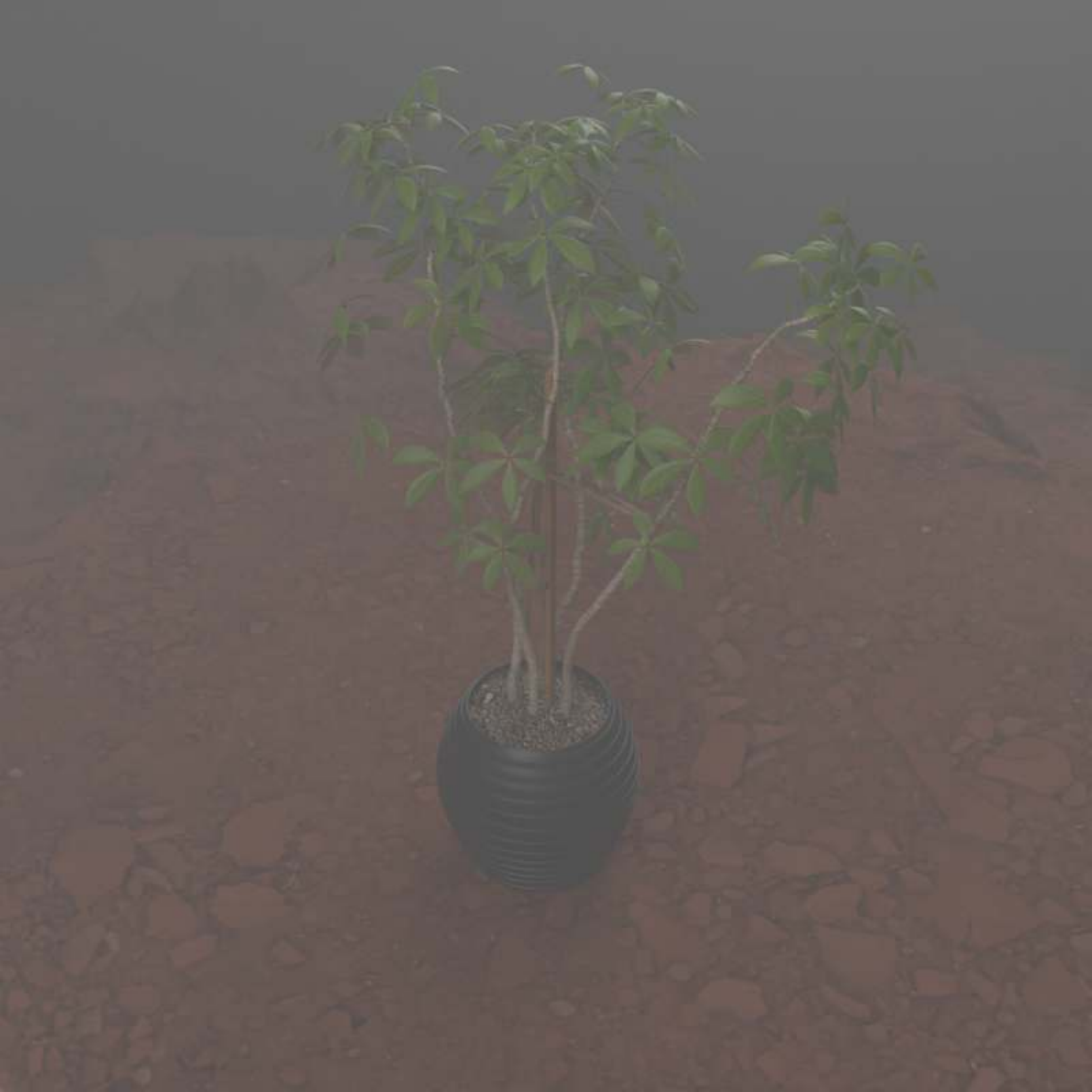}
    \end{subfigure}\hspace{1mm}%
    \begin{subfigure}[]{0.235\textwidth}
        \centering
        \includegraphics[width=0.48\textwidth]{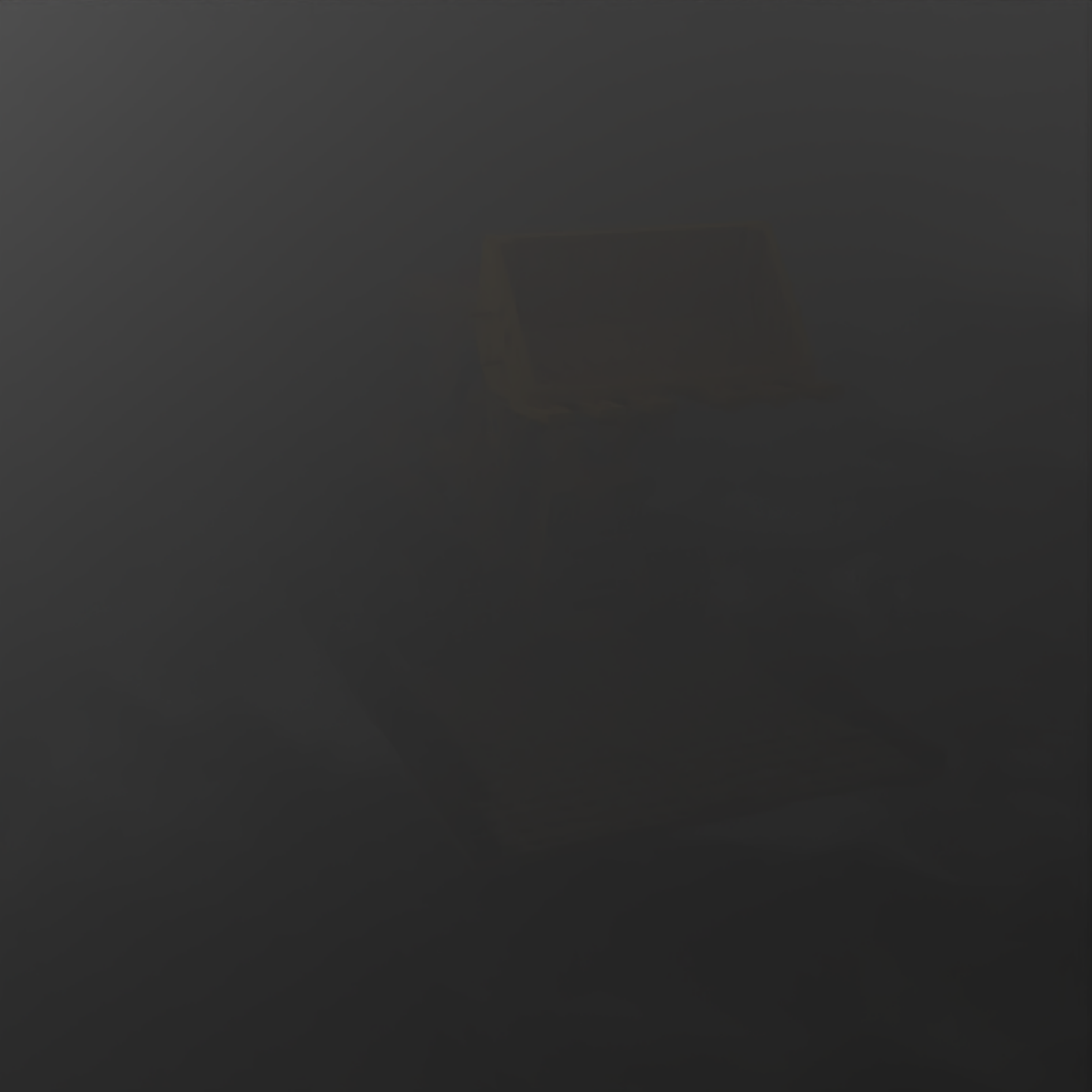}\hspace{1mm}%
        \includegraphics[width=0.48\textwidth]{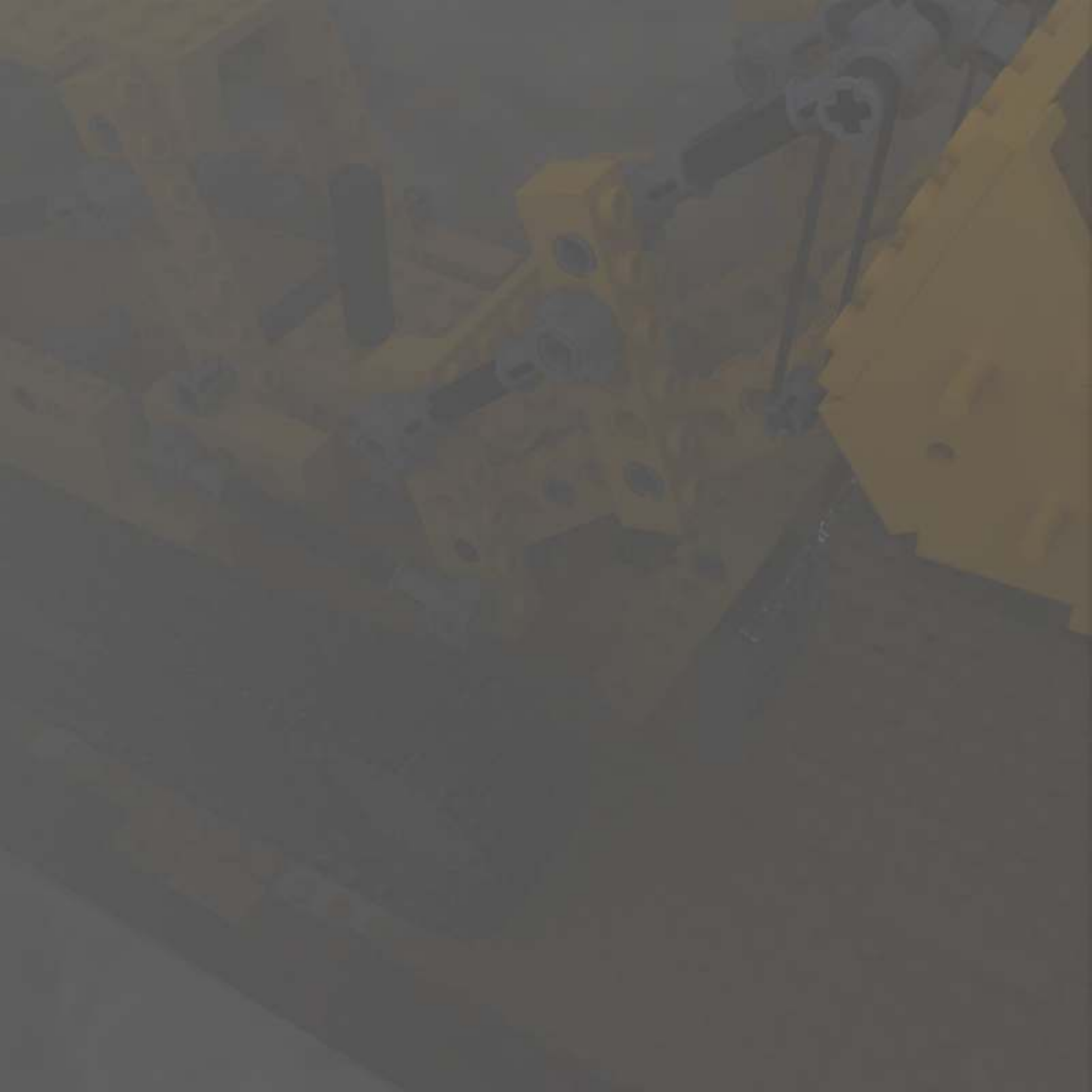}
    \end{subfigure}
    
    \vspace{1mm}

    \rotatebox[origin=c]{90}{\hspace{4mm}NeRF}\hspace{2mm}%
    \begin{subfigure}[]{0.235\textwidth}
        \centering
        \includegraphics[width=0.48\textwidth]{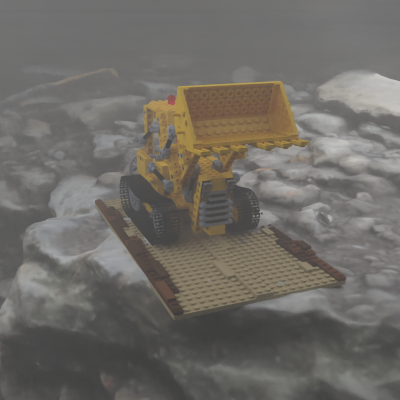}\hspace{1mm}%
        \includegraphics[width=0.48\textwidth]{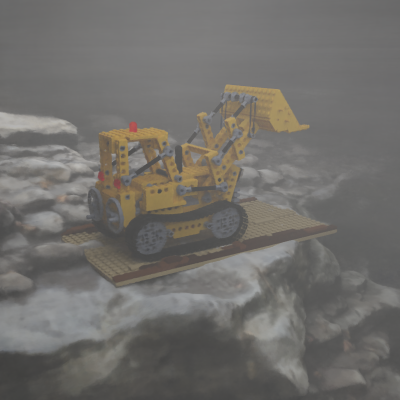}
        \caption{Lego on rocks.}
    \end{subfigure}\hspace{1mm}%
    \begin{subfigure}[]{0.235\textwidth}
        \centering
        \includegraphics[width=0.48\textwidth]{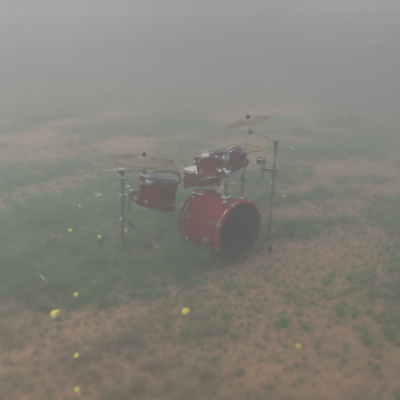}\hspace{1mm}%
        \includegraphics[width=0.48\textwidth]{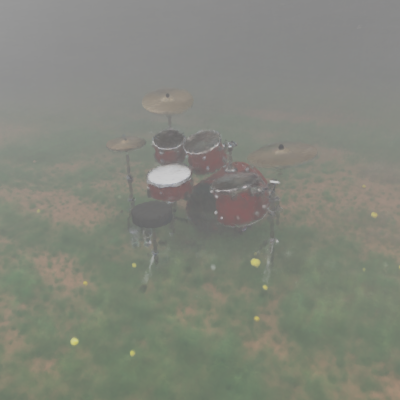}
        \caption{Drums in grassy field.}
    \end{subfigure}\hspace{1mm}%
    \begin{subfigure}[]{0.235\textwidth}
        \centering
        \includegraphics[width=0.48\textwidth]{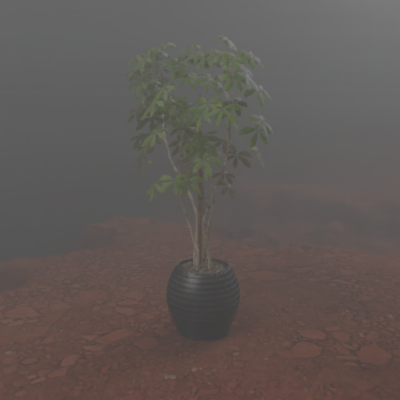}\hspace{1mm}%
        \includegraphics[width=0.48\textwidth]{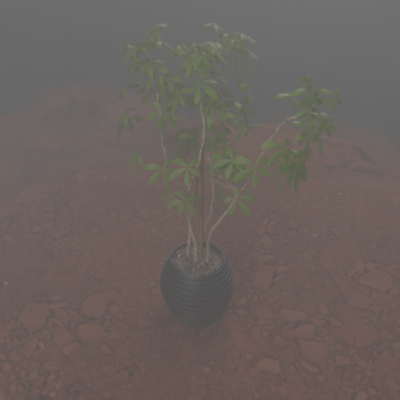}
        \caption{Ficus in desert.}
    \end{subfigure}\hspace{1mm}%
    \begin{subfigure}[]{0.235\textwidth}
        \centering
        \includegraphics[width=0.48\textwidth]{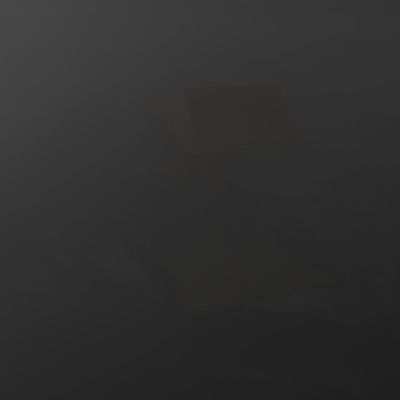}\hspace{1mm}%
        \includegraphics[width=0.48\textwidth]{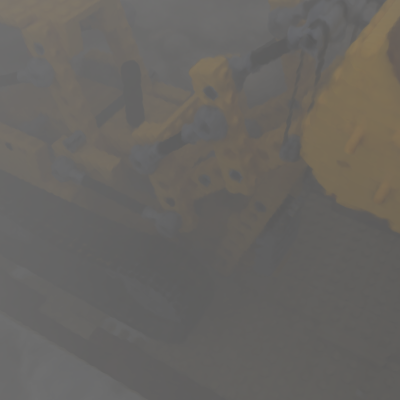}
        \caption{Lego on rocks (heavy fog).}
        \label{subfig:rocks_heavy}
    \end{subfigure}

    \caption{The ground truth images compared with the NeRF synthesized view of the same images from the test set. Note that the dark images on the far right are not an error but rather the effect of heavy fog in these lighting conditions.}
    \label{fig:nerfRenders}
\end{figure*}

\section{Fundamentals of Neural Radiance Fields}

Neural radiance fields tackle the task of synthesizing novel views of complex scenes using a sparse set of images $\mathcal{I}$ with known camera position $\mathbf{X}_{\mathcal{I}}$ and orientation $(\theta, \phi)_{\mathcal{I}}$. This is achieved by representing a radiance field through an MLP with parameters $\mathbf{\theta}$, and continuously optimizing these parameters through a differentiable volume rendering equation. The MLP takes in an encoded 3D positional point $(x,y,z)$ in the radiance field and an encoded view direction $\vec{d}$ and outputs the radiance $\vec{c}$ and density $\sigma$ corresponding to this point. NeRFs render images using ray marching for each pixel ray $\vec{r}(t)=\vec{o} + t\vec{d}$ where the origin $\vec{o}$ and the view direction $\vec{d}$ is derived from $\mathbf{X}_{\mathcal{I}}$ and $(\theta, \phi)_{\mathcal{I}}$. Along the pixel ray $\vec{r}(t)$ a set of radiances and densities are sampled in order to estimate the pixel color $\hat{C}(\vec{r})$ through the volume rendering equation

\begin{equation}
    \label{eq:volrend_anl}
    C(\vec{r}) = \int_{t_n}^{t_f} T(t)\sigma (\vec{r}(t))\vec{c}(\vec{r}(t),\vec{d})\,dt,
\end{equation}

which can be numerically approximated using the quadrature rule from \cite{optmodvolumerend1995}. Let $\delta_i=t_{i+1}-t_{i}$, the distance between samples, resulting in

\begin{equation}
    \label{eq:estimatedcolor}
    \hat{C}(\vec{r}) = \sum_{i=1}^{N} T_i (1-\exp(-\sigma_i \delta_i)) \vec{c}_i.
\end{equation}

$T_i$ is the accumulated transmittance, which denotes the probability that the ray travels from its origin $\vec{o}$ to the point $t_i$ without hitting any dense particles, and is numerically approximated as

\begin{equation}
    \label{eq:transmittanceNum}
    T_i = \exp\left( -\sum_{j=1}^{i-1} \sigma_j\delta_j \right).
\end{equation}

Similarly to the color, the pixel opacity of a ray can be numerically approximated as

\begin{equation}
    \label{eq:ray_opacity}
    \alpha(\vec{r}) = \sum_{i=1}^{N} T_i (1 - \exp(-\sigma_i \delta_i)).
\end{equation}

The optimization of the MLP's parameters $\mathbf{\theta}$ is done through stochastic gradient descent where the photometric loss is calculated between the ground truth colors $C_\text{gt}(\vec{r})$ from $\mathcal{I}$ and the rendered color $\hat{C}(\vec{r})$ using the standard L2 loss function $\mathcal{L} = \sum_{\vec{r}\in \mathcal{R}} \left|\left| \hat{C}(\vec{r}) - C_\text{gt}(\vec{r}) \right|\right|_{2}^{2}$.



\section{Fog Data Generation}
\label{sec:data}

Of the available datasets commonly used for benchmarking NeRF models, including the datasets made available by Mildenhall et al. in the initial NeRF paper~\cite{mildenhall2020nerf}, Tanks and Temples~\cite{Knapitsch2017}, and the Mip-NeRF 360 dataset~\cite{Barron_2022_CVPR}, none contain scenes with any form of fog. Additionally, it is preferable that each scene contains a realistic environment, i.e., not merely a single free-floating object. To accomplish this, we have produced a new synthetic dataset with four scenes. In order to preserve the possibility of some level of comparison with existing NeRF models, the synthetic scenes were created using the same synthetic objects from the dataset introduced by the original NeRF paper.

The first three scenes of our dataset are: Lego bulldozer on rocks, drums in a grassy field, and a ficus plant in a desert canyon. All three were modeled with a moderate amount of fog. Note that the drums in a grassy field scene is purposefully made significantly smaller in scale than the other scenes, thus requiring a higher density of fog for the equivalent appearance. Furthermore, to simulate an extreme scenario, an additional scene of the Lego bulldozer on rocks with heavy fog was created. Each scene is composed of 100 training images with the virtual camera placed in random locations in the upper hemisphere of the scene, pointing towards the center of the object of interest. The random locations are constrained by a distance range from the object and scaled according to the size of the scene. Note that this differs somewhat from the original NeRF dataset \cite{mildenhall2020nerf} by the varying distance from the scene center. We found this to be a vital adjustment to the data-gathering process in order to correctly capture the fog.
These scenes were made as a combination of the synthetic models from \cite{mildenhall2020nerf} and environments published on Blend Swap and TurboSquid~\cite{heinzelnisse2014legobulldozer,bryanajones2014drums,herberhold2019ficus,benthehuman2016rocks,craigforster2016grass,rodox2022desert}. See the first row of fig. \ref{fig:nerfRenders} for example images from the produced dataset.

\begin{figure*}[htbp]
    \centering
    \includegraphics[width=\textwidth]{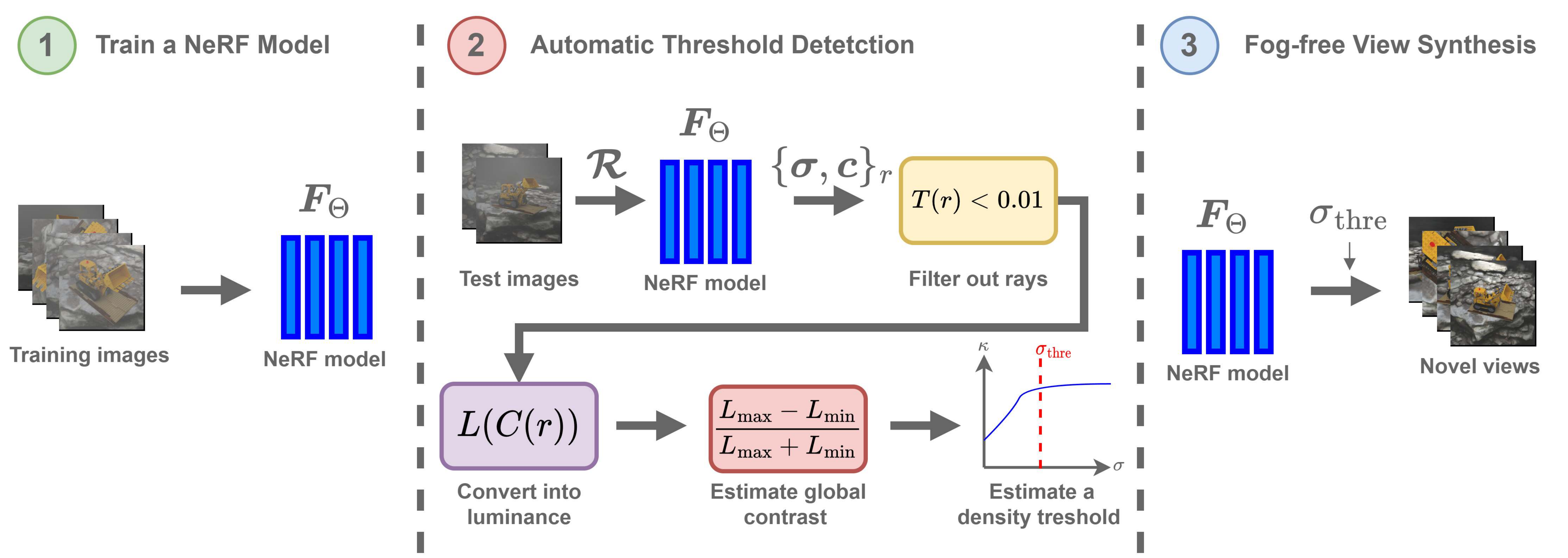}
    \caption{The overview of our model, where a set of pixel rays $\mathcal{R}$ are sampled from a converged NeRF model that has been trained on foggy images. Multiple sets of colors are derived from $\mathcal{R}$ as a function of different density threshold values $\sigma_\text{thre}$ where the colors are then converted to luminance in order to estimate a global contrast of the radiance field. The global contrast as a function of density threshold values $\sigma_\text{thre}$ is used to find the converging point, where this point will be used as a density threshold when synthesizing new views.}
    \label{fig:pipeline}
\end{figure*}

\section{Method}
\label{sec:method}
We propose a novel algorithm for removing volumetric effects, like fog and haze, from generic, pre-trained NeRF models. We do this by applying a density threshold to the model during rendering, ignoring all density values below the threshold. The density threshold is found by optimizing for high, global contrast while keeping a conservative threshold value. An overview of our algorithm can be seen in fig. \ref{fig:pipeline}.

\subsection{Density Threshold for Fog Removal}
\label{sec:density_threshold}

To remove volumetric effects under view synthesis, we apply a density threshold that regulates which samples along a pixel ray $\vec{r}(t)$ will contribute to the final color $\hat{C}(\vec{r})$. Ignoring low-density samples while retaining high-density samples. Our method uses the properties of the volume rendering equation in eq. \ref{eq:estimatedcolor} that forces the radiance field to assign a low density $\sigma_\text{fog}$ for transparent volume. The threshold $\sigma_\text{thre}$ is applied directly on the density $\sigma_i$ output from the MLP, resulting in: 

\begin{equation} 
    \label{eq:depth_threshold_sigma}
    \tilde\sigma_i = \begin{cases}
        \begin{array}{cl}
            \sigma_i & \text{for } \sigma_i \geq \sigma_{\text{thre}} \\
            0 & \text{for } \sigma_i < \sigma_{\text{thre}}.
        \end{array}
    \end{cases}
\end{equation}

The thresholded density value $\tilde\sigma_i$ replaces the original density value when estimating the pixel color through the volume rendering equation, resulting in
\begin{equation}
    \label{eq:estimatedcolorthre}
    \tilde{C}(\vec{r}) = \sum_{i=1}^{N} T_i (1-\exp(-\tilde\sigma_i \delta_i)) \vec{c}_i.
\end{equation}

\begin{figure}
    \centering
    \begin{tikzpicture}[scale=0.75]
        \begin{axis}[
            width=\textwidth,
            xmin=0, xmax=10.4,
            ymin=0, ymax=1.1,
            ticks=none,
            axis lines=left,
            x=1cm, y=4cm,
        ]   
            \addplot[ybar interval,fill=gray!25,draw=gray]coordinates{(0,0.3)(5,0.3)};

            \def \samples {0.0/0.8, 0.25/0.9, 0.5/0.89, 0.75/0.87, 1.0/0.93, 1.25/0.8, 1.5/0.84, 1.75/0.9, 2.0/0.83, 2.25/0.87, 2.5/0.89, 2.75/0.9, 3.0/0.86, 3.25/0.8, 3.5/0.75, 3.75/0.78, 4.0/0.85, 4.25/0.86, 4.5/0.83, 4.75/0.89, 5.0/0.82}

            \foreach \dx / \y in \samples{
                \addplot[ybar interval,fill=blue!25,draw=blue]coordinates{(5+\dx,\y)(5+\dx+0.25,\y)};
            }
        \end{axis}




        \draw [decorate, decoration = {calligraphic brace}, red!50, very thick] (0,0.32*4) --  (5.0, 0.32*4);
        \draw [decorate, decoration = {calligraphic brace}, red!50, very thick] (5.0,0.95*4) --  (10.25, 0.95*4);

        \node[darkgray, above] at (2.5, 0.32*4) {\large \textbf{fog} -- $\sigma_\text{fog}$};
        \node[darkgray, above] at (7.6875, 0.97*4) {\large \textbf{solid}};

        \node[above] at (0, 4.4) {$\sigma$};
        \node[right] at (10.4, 0) {$t$};
    \end{tikzpicture}
    \caption{The densities $\sigma$ along a pixel ray $\vec{r}(t)$ in a theoretical scenario where there is a constant density $\sigma_\text{fog}$ for fog. Due to fog being transparent, its density will be a magnitude lower than the densities associated with solid / non-transparent volume.}
    \label{fig:fog_along_ray}
\end{figure}
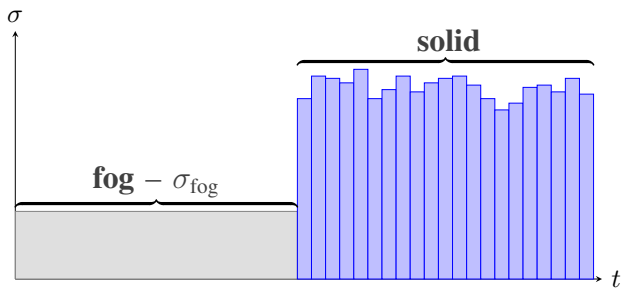

By selecting a low threshold $\sigma_\text{thre}$ we are able to remove the unwanted transparent volume and still keep the solid volume, i.e. objects and surfaces, intact. A theoretical illustration of this is shown in fig. \ref{fig:fog_along_ray} where the fog has a lower density value than the solid part of the radiance field, and therefore an appropriately chosen threshold should remove the fog while leaving the solid object intact. Note that the threshold is applied globally to the whole scene, and is a single value specific to the trained model.

\subsection{Automatic Density Threshold Detection}
\label{sec:auto_dens_thre}

Our method for removing fog under view synthesis requires first identifying a density threshold $\sigma_\text{thre}$ for each scene before view synthesis. 
Although we observed that it is possible to define a constant density threshold for use across several scenes with some success, it will have to be relatively high to cover all cases. This comes with the risk of removing more geometry than desired.

As a result, we propose an automatic process for finding the lowest density threshold that removes fog for each scene individually. This is done by estimating a global contrast of the synthesized images from the radiance field for different values of $\sigma_\text{thre}$, and then choosing the density threshold at the point where there is no longer an increase in contrast. This builds on the assumption that an initial increase in contrast correlates to a decrease in fog. 


\subsection{Estimating Global Contrast}

In order to estimate the global contrast of the synthesized images from the radiance field, we randomly sample a batch $\mathcal{R}$ of pixel rays from any novel view of the model. The density $\sigma$ and radiance $\vec{c}$ value is stored for every sample along all the rays. Next, we select a set of candidate threshold values $Q_\text{thre}$ starting from $0$ up to a theoretical maximum fog density. In our analysis, we find a maximum value of $8$ with candidate values spaced at $0.05$ to be good for most scenes, i.e. $Q_\text{thre} = \left\{0, 0.05, \dots 8\right\}$.

For each value in $Q_\text{thre}$ we apply the thresholding scheme from section \ref{sec:density_threshold} on all samples. Then the final pixel color $\tilde{C}(\vec{r})$ is calculated using the thresholded volume rendering equation eq. \ref{eq:estimatedcolorthre}, for each ray in the batch. 



For each threshold in $Q_\text{thre}$ we calculate the luminances of the pixel colors:

\begin{equation}
    L = 0.2126 \cdot R_\text{linear} + 0.7152 \cdot G_\text{linear} + 0.0722 \cdot B_\text{linear}.
\end{equation}

The maximum and minimum luminance values for the batch of rays associated with each threshold are then used to calculate the Michelson-contrast~\cite{michelson1927studies} as follows:

\begin{equation}
    \label{eq:michcontrast}
    \kappa = \frac{L_\text{max} - L_\text{min}}{L_\text{max} + L_\text{min}}.
\end{equation}

The resulting set of contrast values are ordered by their associated density threshold and smoothed using a Savitzky--Golay filter~\cite{savgol} with polynomial order~$2$ and window size~$21$. These parameters are experimentally tested and chosen to fit the expected curve shape, which exhibits two distinct turning points at the beginning and end of the contrast increase, respectively. The final density threshold value is selected as the first value where the contrast curve flattens out after the initial increase caused by the fog removal. This is determined by the first point where the curve gradient stays at approximately zero for one window length. This value is then used as the global density threshold value for the scene when synthesizing new views. The overall structure of this process and a typical contrast curve can be seen in fig. \ref{fig:pipeline}.

\section{Experiments}
\label{sec:experiments}

We evaluate our proposed algorithm by attempting to synthesize clear views of scenes from our foggy synthetic dataset. All our experiments were conducted using an implementation based on the Python library NerfAcc~\cite{li2022nerfacc}. The generic NeRF models are trained exactly as described in \cite{mildenhall2020nerf}. We only add our density threshold modification to the novel view rendering pipeline after training. 
We let the model train for 1~million steps, where each step consists of forwarding approximately $2^{16}$ samples, i.e. points along the pixel rays, to ensure that the model correctly represents the whole radiance field. 

\subsection{Training NeRF on Foggy Scenes}

We demonstrate NeRF's capacity to capture foggy data in fig. \ref{fig:nerfRenders} together with the corresponding ground truth as rendered by Blender. Visually, we can observe that the standard NeRF model is able to reproduce the scenes to a large extent, but the finer details and edges can become somewhat blurred. 
The dataset's varying distances from the training images to the foggy scene center allow the NeRF model to accurately capture the fog. If the distances from the scene center were fixed, the fog would not be modeled uniformly, thus reducing the accuracy of the generated novel views.
The quantitative results for the three moderately foggy scenes are presented in tab. \ref{tab:detailed_results_fog}. 
These values are significantly higher than the model trained on clear data (tab. \ref{tab:final_result}).  We reason that the fog actually makes the problem easier by removing detail in the background and decreasing the magnitude of the color gradients.

\begin{figure}
    \centering
    \includegraphics{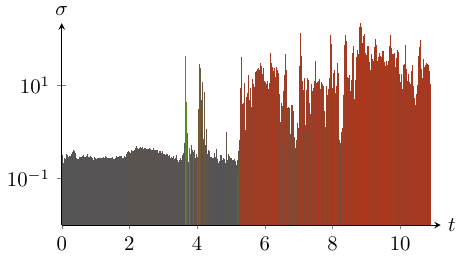}
    \caption{Sample densities along a single pixel ray, corresponding to the central ray of the middle image in fig. \ref{subfig:thresholding_results_desert}.}
    \label{fig:fog_along_ray_ficus}
\end{figure}

For further analysis of how the NeRF model has modeled the radiance field with fog, we experimentally recreate the plot from fig. \ref{fig:fog_along_ray} based on pixel rays that traverse the converged radiance field. This is done by uniform sampling along the rays at set intervals, where each sample will have a density $\sigma_i$ and radiance $\vec{c}_i$. Each sample is plotted as a bar, where density is the height and radiance is the color. The resulting plot for a pixel ray in the ficus in a desert canyon is shown in fig. \ref{fig:fog_along_ray_ficus}. Note that the density is logarithmically scaled. Here, the three elements in the scene are shown: the fog being the gray bars with low-density values, the ficus being the sparse green and brown bars with high density, and the desert environment with the subsurface volume being the orange bars with high density. This distinction between fog and solid volume shows that it is possible to remove fog through thresholding. 

\begin{table}[]
\vspace{2mm}
\centering
\begin{tabular}{|llll|}
\hline
             & PSNR   & SSIM  & LPIPS \\ \hline
Rocks Lego   & 37.076 & 0.948 & 0.248 \\ \hline
Grass Drums  & 33.084 & 0.901 & 0.364 \\ \hline
Desert Ficus & 40.984 & 0.961 & 0.230 \\ \hline
\end{tabular}
    \caption{PSNR, SSIM, and LPIPS scores for different NeRF models after training on the three scenes in moderate fog from our dataset.}
    \label{tab:detailed_results_fog}
\end{table}

\begin{figure*}
    \vspace{2mm}
    \centering

    \begin{subfigure}[]{0.5\linewidth}
        \centering
        \rotatebox[origin=c]{90}{Before}\hspace{4mm}%
        \begin{subfigure}[]{0.3\linewidth}
            \includegraphics[width=\textwidth]{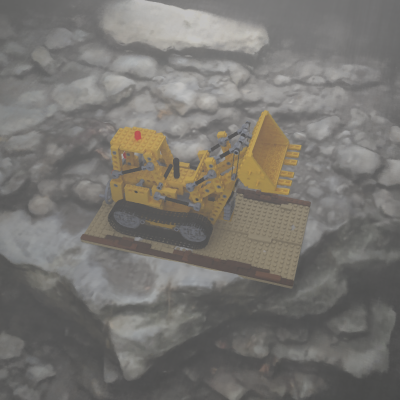}
        \end{subfigure}\hspace{1mm}%
        \begin{subfigure}[]{0.3\linewidth}
            \includegraphics[width=\textwidth]{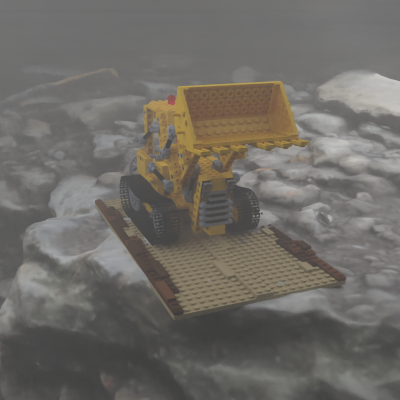}
        \end{subfigure}\hspace{1mm}%
        \begin{subfigure}[]{0.3\linewidth}
            \includegraphics[width=\textwidth]{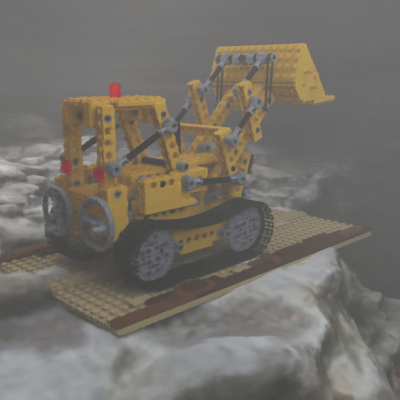}
        \end{subfigure}\\
        \vspace{1mm}
        \rotatebox[origin=c]{90}{After}\hspace{4mm}%
        \begin{subfigure}[]{0.3\linewidth}
            \includegraphics[width=\textwidth]{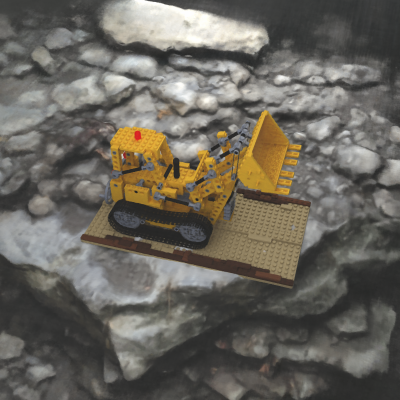}
        \end{subfigure}\hspace{1mm}%
        \begin{subfigure}[]{0.3\linewidth}
            \includegraphics[width=\textwidth]{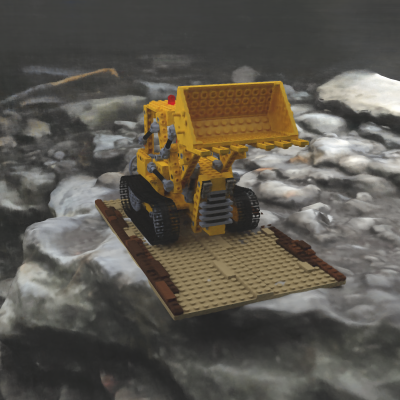}
        \end{subfigure}\hspace{1mm}%
        \begin{subfigure}[]{0.3\linewidth}
            \includegraphics[width=\textwidth]{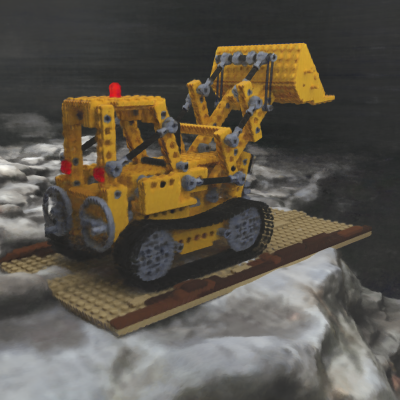}
        \end{subfigure}
        \caption{Lego bulldozer on rocks in moderate fog with $\sigma_\text{thre} = 0.75$.}
        \label{subfig:thresholding_results_rocks}
    \end{subfigure}%
    \begin{subfigure}[]{0.5\linewidth}
        \centering
        \begin{subfigure}[]{0.3\linewidth}
            \includegraphics[width=\textwidth]{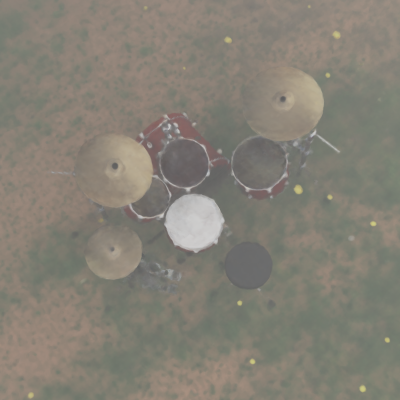}
        \end{subfigure}\hspace{1mm}%
        \begin{subfigure}[]{0.3\linewidth}
            \includegraphics[width=\textwidth]{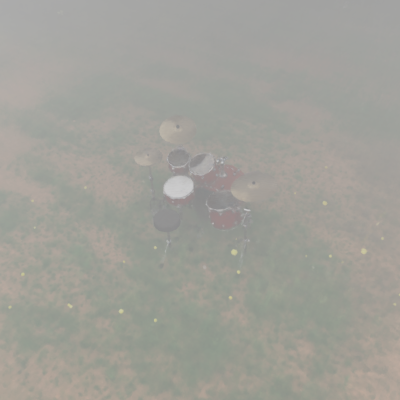}
        \end{subfigure}\hspace{1mm}%
        \begin{subfigure}[]{0.3\linewidth}
            \includegraphics[width=\textwidth]{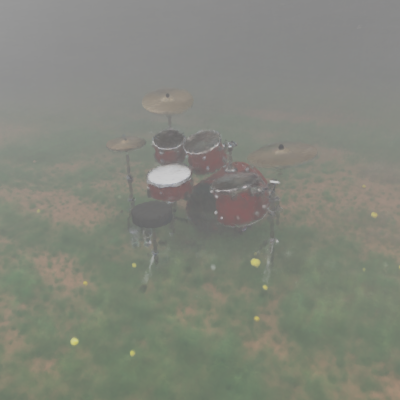}
        \end{subfigure}\\
        \vspace{1mm}
        \begin{subfigure}[]{0.3\linewidth}
            \includegraphics[width=\textwidth]{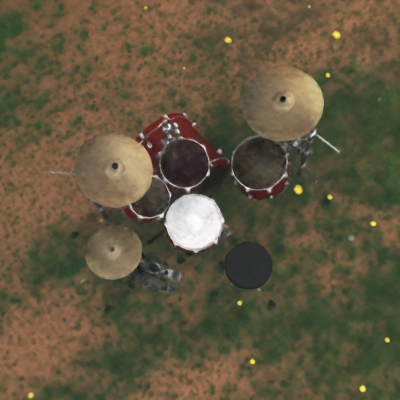}
        \end{subfigure}\hspace{1mm}%
        \begin{subfigure}[]{0.3\linewidth}
            \includegraphics[width=\textwidth]{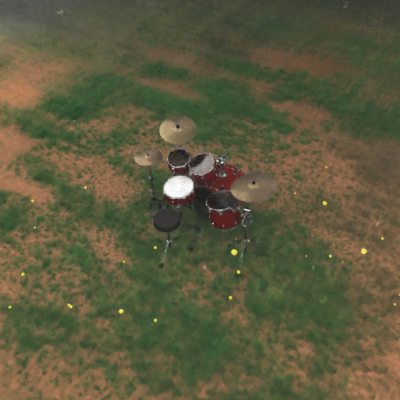}
        \end{subfigure}\hspace{1mm}%
        \begin{subfigure}[]{0.3\linewidth}
            \includegraphics[width=\textwidth]{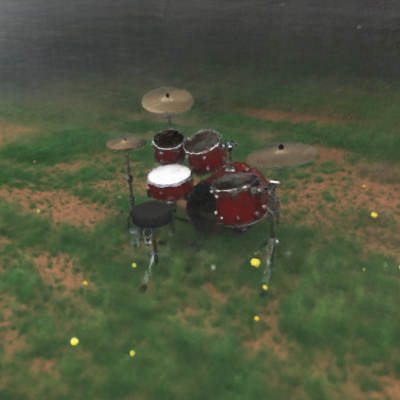}
        \end{subfigure}
        \caption{Drums in a grassy field with $\sigma_\text{thre}=3.00$.}
        \label{subfig:thresholding_results_grass}
    \end{subfigure}

    \vspace{2mm}
    
    \begin{subfigure}[]{0.5\linewidth}
        \centering
        \rotatebox[origin=c]{90}{Before}\hspace{4mm}%
        \begin{subfigure}[]{0.3\linewidth}
            \includegraphics[width=\textwidth]{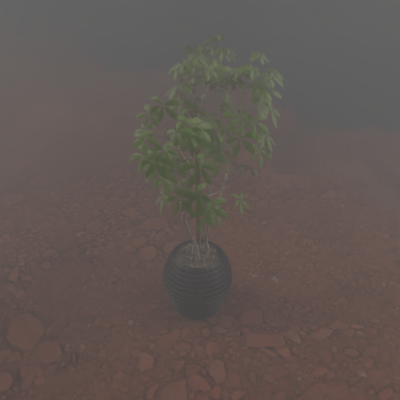}
        \end{subfigure}\hspace{1mm}%
        \begin{subfigure}[]{0.3\linewidth}
            \includegraphics[width=\textwidth]{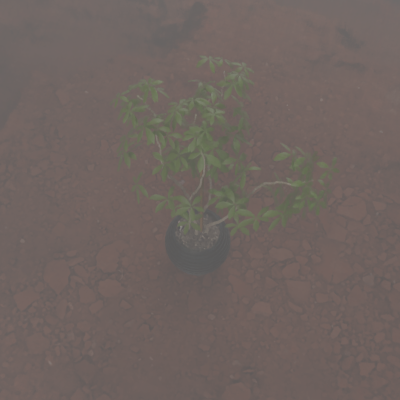}
        \end{subfigure}\hspace{1mm}%
        \begin{subfigure}[]{0.3\linewidth}
            \includegraphics[width=\textwidth]{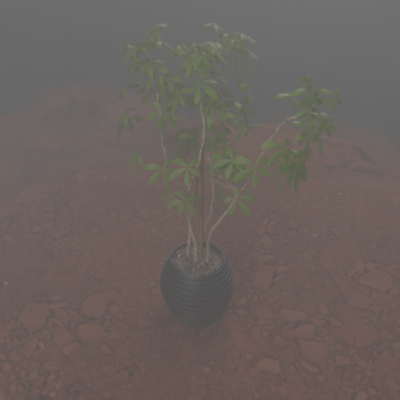}
        \end{subfigure}\\
        \vspace{1mm}
        \rotatebox[origin=c]{90}{After}\hspace{4mm}%
        \begin{subfigure}[]{0.3\linewidth}
            \includegraphics[width=\textwidth]{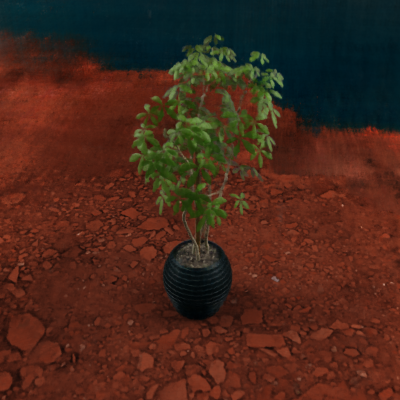}
        \end{subfigure}\hspace{1mm}%
        \begin{subfigure}[]{0.3\linewidth}
            \includegraphics[width=\textwidth]{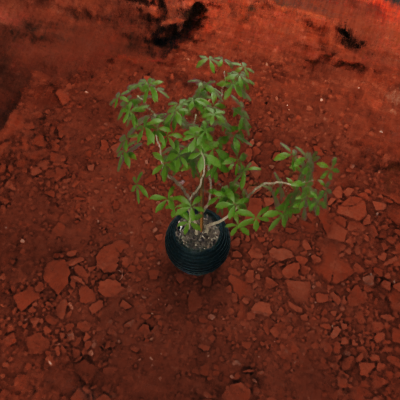}
        \end{subfigure}\hspace{1mm}%
        \begin{subfigure}[]{0.3\linewidth}
            \includegraphics[width=\textwidth]{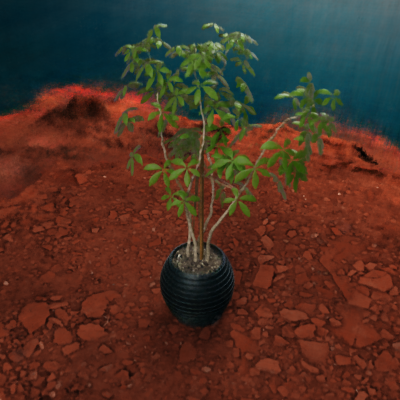}
        \end{subfigure}
        \caption{Ficus plant in a desert canyon with $\sigma_\text{thre}=0.95$.}
        \label{subfig:thresholding_results_desert}
    \end{subfigure}%
    \begin{subfigure}[]{0.5\linewidth}
        \centering
        \begin{subfigure}[]{0.3\linewidth}
            \includegraphics[width=\textwidth]{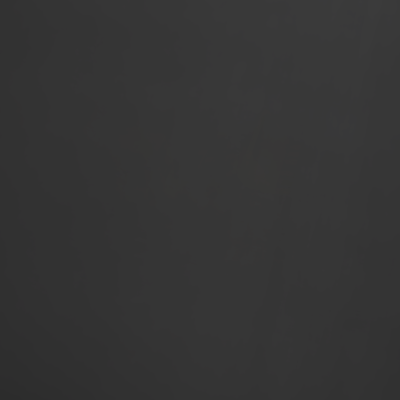}
        \end{subfigure}\hspace{1mm}%
        \begin{subfigure}[]{0.3\linewidth}
            \includegraphics[width=\textwidth]{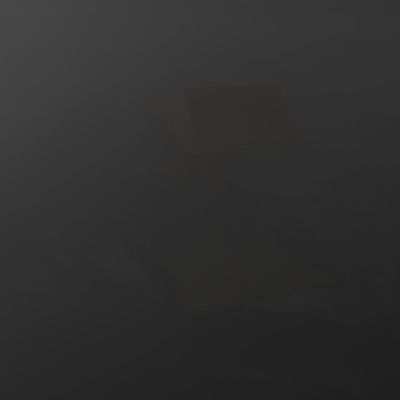}
        \end{subfigure}\hspace{1mm}%
        \begin{subfigure}[]{0.3\linewidth}
            \includegraphics[width=\textwidth]{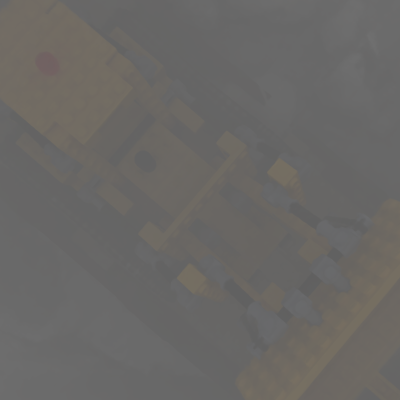}
        \end{subfigure}\\
        \vspace{1mm}
        \begin{subfigure}[]{0.3\linewidth}
            \includegraphics[width=\textwidth]{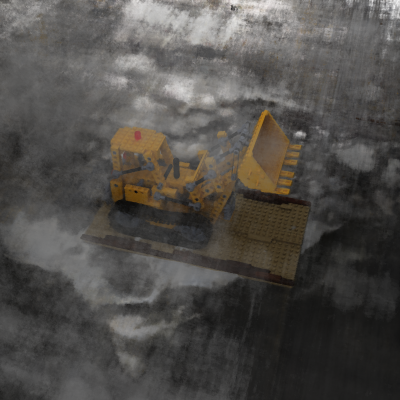}
        \end{subfigure}\hspace{1mm}%
        \begin{subfigure}[]{0.3\linewidth}
            \includegraphics[width=\textwidth]{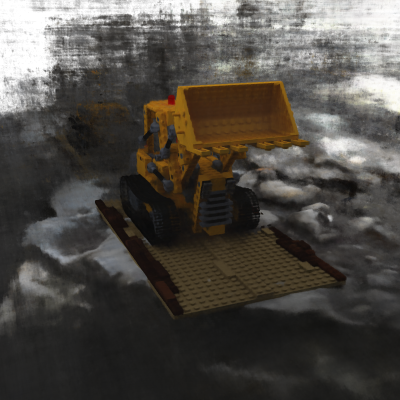}
        \end{subfigure}\hspace{1mm}%
        \begin{subfigure}[]{0.3\linewidth}
            \includegraphics[width=\textwidth]{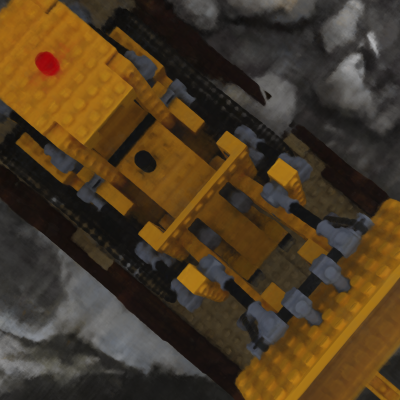}
        \end{subfigure}
        \caption{Lego bulldozer on rocks in heavy fog with $\sigma_\text{thre} = 1.90$.}
        \label{subfig:thresholding_results_heavy}
    \end{subfigure}

    \caption{Novel view synthesis before and after applying our automatic density thresholding.}
    \label{fig:thresholding_results}
\end{figure*}

\subsection{Removing Fog from a Trained NeRF Model}
\begin{table}[]
\centering
\begin{tabular}{|llll|}
\hline
  & Rocks Lego & Grass Drum & Desert Ficus \\ \hline
Clear & 29.16 & 24.46 & 27.34 \\ \hline
Our & 22.55 & 21.84 & 18.75 \\ \hline
He et al. \cite{he2010single} & 17.60 & 20.42 & 17.97 \\ \hline
\end{tabular}
    \caption{Indicative, quantitative result of fog removal compared to NeRF trained on the clear dataset and a seminal paper on fog removal.}
    \label{tab:final_result}
\end{table}

With a pre-train NeRF model for each of the four foggy scenes, we apply our automatic threshold detection scheme from section \ref{sec:auto_dens_thre} in order to determine a global density $\sigma_\text{thre}$ threshold. The density thresholds for a given scene are then used to synthesize novel views without fog. Fig. \ref{fig:thresholding_results} shows the density thresholds for the different scenes in our dataset, with example renderings before and after removing the fog. Subfigures \ref{subfig:thresholding_results_rocks}, \ref{subfig:thresholding_results_grass}, and \ref{subfig:thresholding_results_desert} show the results of the scenes with moderate fog. The effects of fog are greatly reduced after applying our thresholding scheme, leaving the scene center free of fog. For some images like \ref{subfig:thresholding_results_desert} (right), we can see some background coloration not present in the original dataset. This is due to the fog being heavy enough that the background is not visible during training, resulting in the absence of the background in the rendered images. Also note that the selected threshold for the drums in a grassy field scene is significantly higher than the other scenes with a similar apparent amount of fog due to the smaller scale of the scene as discussed in section \ref{sec:data}.

To push the limits of our method, we have also performed one test with very heavy fog (fig. \ref{subfig:thresholding_results_heavy}), where the center Lego truck is almost completely occluded, even at relatively short distances. But despite this extreme situation, our method still manages to remove all fog volume for close views and significantly reduce fog-related volume for far views.
However, the edges of the radiance field, which has had little coverage in the training images, show a gradual increase in fog-like volume. This is due to the volume-constrained radiance fields, where the edges would have to contain enough dense volume to "paint" the scene outside the field.
Both the "tinting effect" and the "painting effect" might be further removed by ignoring the edges of the volume when rendering, although some edge geometry will also have to be discarded.

Due to the difference in lighting between the clear dataset and the images with fog removed as mentioned in \ref{sec:relatedwork}, it is not easy to establish a metric to verify the effectiveness of fog removal. But to give some quantitative results, we have attempted to create an algorithm based on the PSNR score to give some sense of performance. To avoid the problems with unobserved details, we remove all pixels with a ground truth depth value above some threshold. We set this threshold such that roughly 50\% of the image pixels are included. 
Then, we check the color/illumination-dependent PSNR score between the rendered image and the ground truth. This is done for each image individually before taking the averages for each scene. This value gives a lower bound performance metric for our method as it tries to compare two images under different lighting conditions. Tab \ref{tab:final_result} shows the results for a NeRF model trained on clear data, our method after fog removal, and a seminal paper in single image dehazing for reference. Although the PSNR drops somewhat between the NeRF model trained in clear conditions and our method after fog removal, the PSNR score is still decent, and our method beats the single image dehazer for all scenes.

\section{Conclusion}
\label{sec:conclusion}

We have presented a general method extending NeRFs to allow fog-free novel view synthesis of models trained on foggy images. This was done by changing the volume rendering equation to disregard the fog volume. To determine whether the volume is fog or not, we argue that low-density values are a sufficient discriminator and that it can be thresholded away during view synthesis. In order to derive a threshold value for density, we propose a method that estimates a global contrast as a function of density thresholds and use the convergence point as a global density threshold to be applied to the entire radiance field. Our experiments show that our method can synthesize fog-free novel views from NeRFs trained on images depicting foggy environments. The simplicity of our method allows it to be extended on top of generic NeRF models. We hope our findings will help extend future NeRF-based models to work in more adverse environments.






\bibliographystyle{IEEEtran}
\bibliography{IEEEabrv,egbib}

\end{document}